\documentclass[twocolumn]{jov}

\usepackage{float}
\usepackage{hyperref}
\usepackage{fancyhdr}
\usepackage{amsmath}
\usepackage{amsfonts}
\usepackage{amssymb}
\usepackage{amscd}
\usepackage{latexsym}
\usepackage{graphicx}
\usepackage{subcaption}

\usepackage{gensymb}
\begin{document}
	
\twocolumn[
\begin{@twocolumnfalse}
	
\title{Deep Neural Models for color discrimination and color constancy}

\abstract{
Color constancy is our ability to perceive constant colors across varying illuminations. Here, we trained deep neural networks to be color constant and evaluated their performance with varying cues. Inputs to the networks consisted of the cone excitations in 3D-rendered images of 2115 different 3D-shapes, with spectral reflectances of 1600 different Munsell chips, illuminated under 278 different natural illuminations. The models were trained to classify the reflectance of the objects. One network, Deep65, was trained under a fixed daylight D65 illumination, while DeepCC was trained under varying illuminations. Testing was done with 4 new illuminations with equally spaced CIEL*a*b* chromaticities, 2 along the daylight locus and 2 orthogonal to it. We found a high degree of color constancy for DeepCC, and constancy was higher along the daylight locus. When gradually removing cues from the scene, constancy decreased. High levels of color constancy were achieved with different DNN architectures. Both ResNets and classical ConvNets of varying degrees of complexity performed well. However, DeepCC, a convolutional network, represented colors along the 3 color dimensions of human color vision, while ResNets showed a more complex representation. }

\author{Flachot}{Alban}
 {Abteilung Allgemeine Psychologie}
 {Justus Liebig University Giessen}
 {}
 {Alban.Flachot@psychol.uni-giessen.de}
 
 \author{Akbarinia}{Arash}
 {Abteilung Allgemeine Psychologie}
 {Justus Liebig University Giessen}
 {}
 {Arash.Akbarinia@psychol.uni-giessen.de}

\author{Sch\"utt}{Heiko H.}
 {Center for Neural Science}
 {New York University}
 {}
 {heiko.schuett@nyu.edu}
 
\author{Fleming}{Roland W.}
 {Experimental Psychology}
 {Justus Liebig University Giessen}
 {}
 {}
 
\author{Wichmann}{Felix A.}
 {Neural Information Processing Group}
 {University of T\"ubingen}
 {}
 {}
 
\author{Gegenfurtner}{Karl R.}
 {Abteilung Allgemeine Psychologie}
 {Justus Liebig University Giessen}
 {}
 {gegenfurtner@uni-giessen.de}


\maketitle
\end{@twocolumnfalse}
]

\section{Introduction} \label{sec:introduction}

Color constancy denotes the ability to perceive constant colors, even though variations in illumination change the spectrum of the light entering the eye. Although extensively studied (see \cite{gegenfurtner2003color, witzel2018color, Foster:2011} for reviews), it has yet to be fully understood. Behavioral studies disagree on the degree of color constancy exhibited by human observers \cite{witzel2018color} and color constancy is considered an ill-posed problem. It is argued from theoretical and mathematical considerations that perfect color constancy is not possible using only the available visual information \cite{maloney1986color, logvinenko2015rethinking}. Yet, observing that humans do achieve at least partial color constancy sparks the question about which cues and computations they use to do so. It also remains unclear which neural mechanisms contribute to color constancy. Low-level, feedforward processes, such as adaptation and the double opponency of cells in early stages of the visual system, have been identified as being useful for color constancy \cite{gao2015color}. Yet, other studies suggest that higher-level and even cognitive processes such as memory also contribute. For example, better color constancy has been observed for known objects than for unknown ones \cite{granzier2012effects, olkkonen2008color}. 
A complete neural model of color constancy, both encompassing physiological properties similar to the primate's visual system, and at the same time exhibiting similar behaviour to humans on color constancy relevant tasks, thus remains to be developed. 

Color constancy is also a well studied problem in computer vision and image processing, yet the extent to which the algorithms in these engineering fields can inform our understanding of human color constancy is limited. In those fields, it is often defined by the estimation of the scene's illumination \cite{land1964retinex, akbarinia2017colour} followed by an image correction via the Von Kries assumption \cite{von1902chromatic}. This is in contrast to biological vision, where color constancy is rather defined as the ability to extract color information about the object and materials in the scene consistently across varying illuminations \cite{maloney1986color,Foster:2011, witzel2018color}. Moreover, computer vision studies are typically performance focused: they generally do not investigate \textit{how} models estimate and represent the illumination in a picture, it is enough that the illumination is estimated accurately. Thus their focus is typically on image processing rather than on understanding human color vision.

In contrast to earlier computer vision approaches, Deep Neural Networks (DNNs) may have greater potential to be models for biological color constancy and color vision.
Conceptually inspired by biology \cite{lecun1995convolutional}, DNNs can solve many complex visual tasks such as face and object recognition \cite{zeiler2014visualizing, yosinski2015understanding}, and DNNs trained for object recognition have been shown to correlate with neuronal activity in visual cortical regions \cite{gucclu2015deep,cichy2016deep}. The predictions for cortical activity are not perfect though, and DNN responses are far less robust to distortions of the input images than human observers \cite{goodfellow2014explaining, brendel2017decision, geirhos2017comparing, akbarinia2020deciphering}. Furthermore, it has been shown that current DNNs and human observers do not agree which individual images they find easy or difficult to recognize \cite{Geirhos_2020b}.

For the processing of color information specifically, similarities have been observed between DNNs trained on complex tasks and the visual system \cite{rafegas2018color, flachot2018processing}. In addition, DNNs trained on illumination estimation from images have outperformed all previous approaches \cite{lou2015color, bianco2015color, hu2017fc4, shi2016deep, afifi2019sensor}. This success was enabled by fine tuning networks pretrained on other tasks \cite{lou2015color}, various data augmentation techniques including the application of additional color distortions and cropping \cite{lou2015color, bianco2015color}, and architectural innovations and adversarial training \cite{hu2017fc4, shi2016deep, afifi2019sensor}. Notably, none of these networks were trained only on natural variation in illuminations, and most of them still aimed at the task of color correcting images, not at estimating object color.


Deep learning approaches to color constancy are limited by their need for large datasets. The heavy requirements for a good color constancy image dataset (calibrated cameras, pictures taken from the same angle at different times of day, or with many different controlled and measured illuminations) result in datasets rarely containing more than a thousand images \footnote{see \url{https://colorconstancy.com/evaluation/datasets/} for a review}. One approach to generate larger training datasets for this kind of situation is to use computer graphics to render images or videos instead. This approach has successfully been used for depth and optical flow estimation tasks \cite{butler2012naturalistic, dosovitskiy2015FlowNet, ilg2018Occlusions}, but has to our knowledge not been applied to color constancy yet.



To train networks only on color constancy and make them more comparable to human color vision, we proceeded as follows: Where previous approaches overcame the limited set for training images through data-augmentation, we generated artificial training and validation images using 3D spectral rendering with a naturalistic distribution of illuminations. Instead of RGB encoded inputs, we used images encoded using human cone sensitivities. 
Instead of training our models on illumination estimation, we trained them to extract object colors within the scene. Specifically, the task was to classify objects floating in a room based on their surface color, under a large set of different illumination conditions. Chromaticities of colored surfaces and illuminations were such that color constancy was necessary to attain high accuracy. We then devised an evaluation procedure of the trained models to allow comparison with human studies. Finally, instead of using only a large, complicated standard deep learning model, we trained both complex and relatively simple ones, and compared their performance as well as the color representations they develop during training. 

We found that all our models performed very well at recognizing objects surface colors, even for illuminations they had never seen, with a supra-human accuracy. 
Like humans \cite{kraft1999mechanisms}, the accuracy of the models drastically degrades, however, as we manipulate the input by gradually removing cues necessary for color constancy. Similarly, we also found a better performance for illuminations falling along the daylight axis than for illuminations falling in the orthogonal direction. This result is in line with observations made in psychophysical studies \cite{pearce2014chromatic,aston2019illumination}. We found, however, that different architectures learned to represent the surface colors of objects very differently. One of them, \textit{DeepCC}, the most straightforward convolutional architecture we implemented, seems to represent surface colors following criteria resembling the perceptual color dimensions of humans, as determined by psychophysical studies. Other architectures like ResNets, on the other hand, did not.  This suggests that while perceptual color spaces may aid color constancy, they are certainly not necessary for achieving human-like robustness to changes in illumination.

This manuscript is divided into sections following our main findings. We start by reporting the results obtained for DeepCC's evaluation, with a focus on the effect of illumination on DeepCC's performance. Then we analyse how DeepCC represents surface colors and gradually becomes color constant throughout its processing stages. We finish with a summary of the results obtained for other deep net architectures, in particular custom ResNet architectures.

\section{General Methods} \label{sec:methods}

\subsection{Munsell and CIEL*a*b* coordinates}\label{ssec:color spaces}

Throughout this study, 2 color coordinate systems are used.
The first one is the Munsell color system \cite{munsell1912pigment, cleland1921practical}, defined by the Munsell chips themselves. Each Munsell chip is indexed according to 3 coordinates: \textit{Hue}, \textit{Value} and \textit{Chroma}. \textit{Hue} is divided into 5 main hues: Red, Yellow, Green, Blue and Purple, each one divided into 8 intermediary hues, for a total of 40 hues. \textit{Value} is close to \textit{lightness} as it refers to how light a Munsell chip is perceived to be. In terms of surface reflectance, it approximately corresponds to the amount of light that gets reflected by the Munsell chip, i.e., the area under curve \cite{flachot2019extensions}. \textit{Value} varies from 0 to 10, 0 being the darkest and 10 being the lightest. \textit{Chroma} refers to the colorfulness of the chip, or its distance from grey. In terms of surface reflectance, it corresponds to the contrast in the amount of light reflected by different wavelengths. The higher the chroma, the less flat the surface reflectance spectrum \cite{flachot2019extensions} and the more colorful the chip. \textit{Chroma} varies from 0 to 16. Note however that the Munsell color system does not have perfect cylindrical shape but has a limited gamut: certain hues and values do not allow for high chromas. Hence the full set of Munsell chips consists of only 1600 chips instead of $40 \times 16 \times 10 = 5600$ chips. Because the Munsell color system is defined by the Munsell chips, it is the most appropriate space to discriminate Munsells. In addition, the Munsell chips were chosen in an attempt be perceptually uniformly distant and as such, the Munsell coordinate system is an approximately perceptually uniform space.

Another perceptually uniform color space is the CIEL*a*b* \cite{CIELAB} coordinate system.  It was constructed such that its euclidean distance, commonly called $\Delta$ E, is an approximate measure of perceptual difference: two colors equidistant to another in CIEL*a*b* are \textit{approximately} perceptually equidistant. Of the three dimensions, \textit{L*} accounts for lightness, \textit{a*} accounts for greenish-reddish variations and b* accounts for blueish-yellowish variations. The white point (point of highest Lightness) was computed using the spectrum of the light reflected by the Munsell Chip of highest value, under the D65 illumination. This Munsell chip is also an achromatic chip.


\subsection{Image generation}\label{ssec:renderings}

In the present study, we generated our own images using the physically based renderer Mitsuba \cite{jakob2010mitsuba}. Mitsuba was developed for research in physics and includes accurate, physics-based approximations for the interaction of light with surfaces \cite{pharr2016physically, bergmann2016phenomenological} yielding a perceptually accurate rendering \cite{guarnera2018perceptually}. Most importantly, it also allows the use and rendering of spectral data: one can use physically measured spectra of lights and surfaces as parameters. Outputs can also be multi-spectral images rather than simple RGB images. We exploited this multi-spectral characteristic of Mitsuba using the reflectance spectra of 1600 Munsell chips \cite{munsell1912pigment} downloaded from Joensuu University \footnote{\url{http://www.cs.joensuu.fi/\~spectral/databases/download/munsell\_spec\_glossy\_all.htm}} \cite{orava2003color}. 
As illuminations, we used the power spectra of 279 natural lights: 43 were generated from the D series of CIE standard illuminations \cite{judd1964spectral,schanda2007colorimetry} at temperatures ranging from 4.000K to 12.000K; 236 were taken from the forest illuminations measured by Chiao et al. \cite{chiao2000color}. Each illumination spectrum was normalized such that their highest point reaches the same, arbitrary value of a 100.

For meshes, we used a compilation of object datasets issued by Evermotion \footnote{\url{https://evermotion.org/shop}}, for a total of 2115 different meshes, ranging from man-made objects to natural objects. Each mesh was normalized such that they have the same size (equal longuest dimension). 

\begin{figure*}[h]
    \centering
		\begin{tabular}{@{\hskip5pt}c}
			 	\includegraphics[width=1.95\columnwidth]{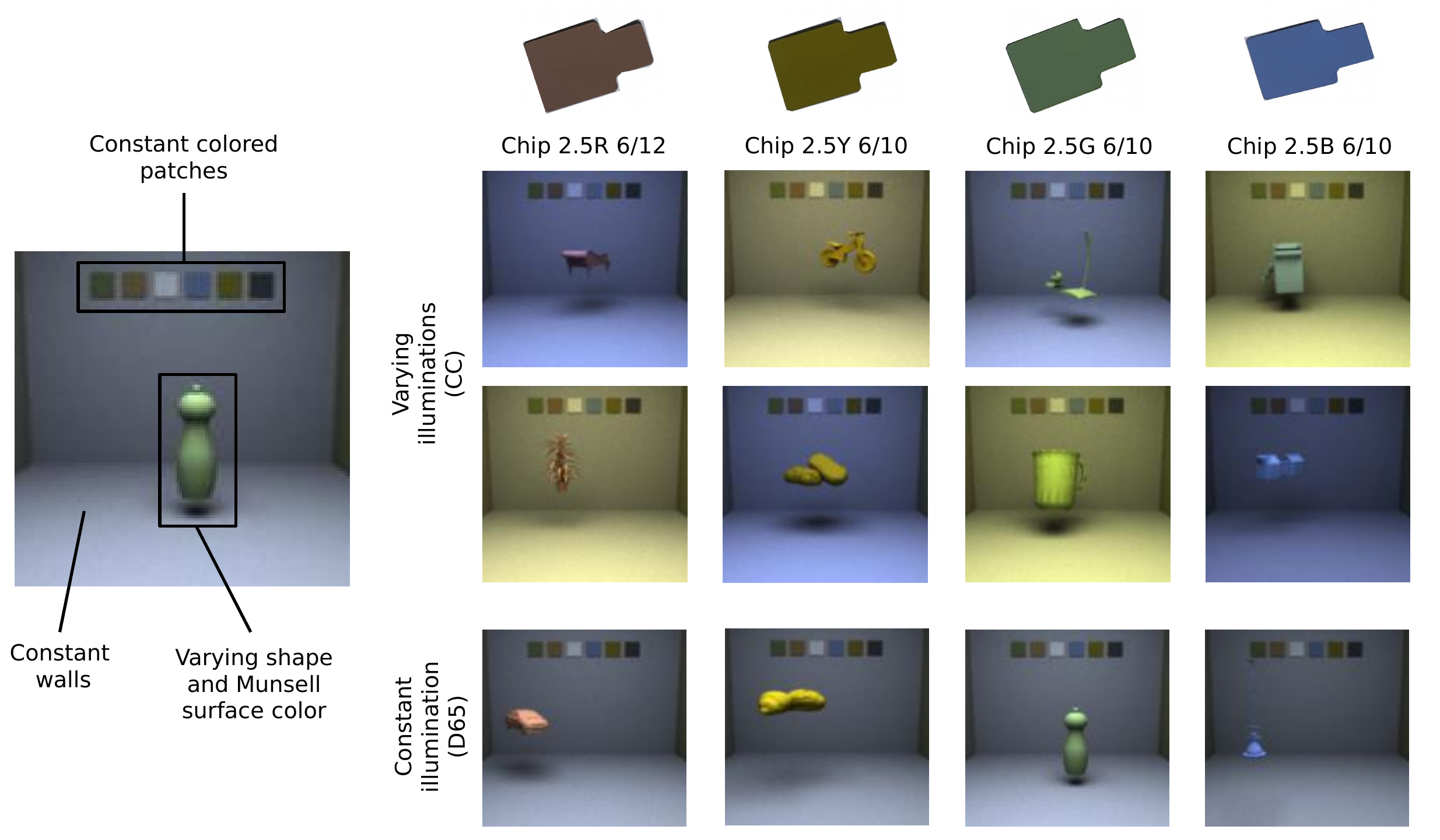}
        \end{tabular}
        
		\caption{Illustration of the 2 training datasets used: one with varying illumination (CC), another with a constant illumination (D65). The classification task consisted of identifying the correct Munsell chip used as surface reflectance for a random object floating in the room. Note that to be performant on the CC dataset, the network had to account for the illumination.}
		\label{fig:example_sets}
\end{figure*}

In order to approximate the input to human visual processing, we first generated our images with 20 channels, at equally-spaced wavelengths ranging from 380 to 830 nm. These were then collapsed onto 3 channels using measured human cone sensitivities \cite{stockman2000spectral}. Images were saved with floating points, thus without the need for any gamut correction or further processing. 
The 3D scene consisted of a simple ``room" (see Figure \ref{fig:example_sets}), with 3 walls, a floor and a ceiling with constant Munsell reflectances as surfaces. On the ceiling, a rectangular light source was defined. On the back wall, 6 colorful patches with constant Munsell reflectances were added. Their purpose was giving additional cues for the model to solve color constancy, as seems to be necessary for humans \cite{brainard2003color, yang2001illuminant}. 

Finally, each LMS image consisted of a random object floating at a random position and orientation in the scene, with a given Munsell surface reflectance. We generated 2 datasets, the \textit{Set-CC} and \textit{Set-D65} datasets. Illustrations of these datasets are available in Figure \ref{fig:example_sets}. In the \textit{CC} dataset, we generated 279 images per Munsell chip, one for each of the 279 natural illuminations. In the \textit{D65} dataset, we also generated 279 images per Munsell chip value, but kept the illumination constant with the power spectrum of the standard D65 illumination. Each dataset thus consisted of $1600 \times 279 = 446400$ images. Images were labeled according to the mesh type, object position, illumination and most importantly in this study, according to the Munsell chip used for the mesh's surface reflectance.
All surfaces were defined as Lambertian.
This dataset is publicly available from [link provided upon acceptance], as well as the pipeline to generate it.

\subsection{Deep architecture}\label{ssec:models}

One network architecture has been extensively studied throughout this work. Several others were also tested, evaluated and analyzed, for which results are described in detail in Section \textit{Standard and custom architectures}. For now, we limit ourselves to describing the network architecture most relevant for this study, which we refer to as \textit{Deep}. 

\textit{Deep} has a convolutional architecture \cite{lecun1998gradient, krizhevsky2012imagenet} with 3 convolutional layers and 2 fully connected layers preceding a classification layer. Convolutional layers can be described as a set of linear kernels. Each kernel applies the same linear filter of limited size on different portions of the input, at regular intervals. The output of one linear filter applied on one input patch, coupled with a half-wave rectification (ReLU), is the output of one unit. Units in convolutional layers have thus limited receptive fields in the image input. Fully connected layers instead take all units of the previous layer as input, such that the units' receptive fields cover the whole input image. The layers of the \textit{Deep} architecture have 16, 32, 64, 250 and 250 kernels respectively. The kernel sizes of the 3 convolutional layers are subsequently 5, 3 and 3. After each convolutional layer is a 2$\times$2 maxpooling layer. The classification layer is a simple fully connected layer, preceded by a 40\% dropout layer for regularization.

\subsection{Task and training}\label{ssec:training}

\begin{figure*}[h]
    \centering
		\begin{tabular}{@{\hskip5pt}c}
			 	\includegraphics[width=1.95\columnwidth]{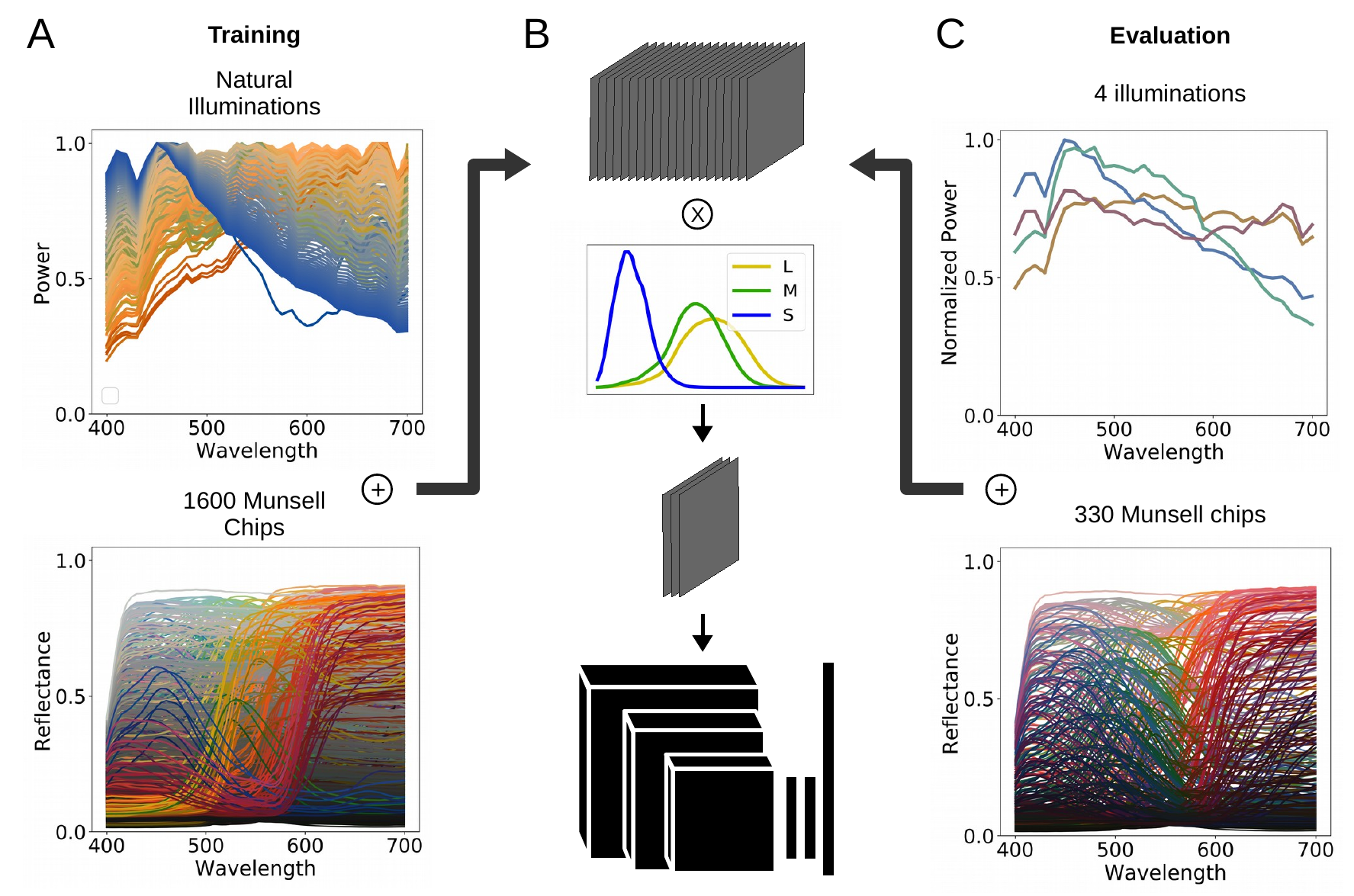}
        \end{tabular}
        
		\caption{Figure illustrating our method, both for training and evaluation. (\textit{Panel A}:) To generate the training set of images, sets of 279 spectra of natural illuminations and 1600 spectra of Munsell reflectances were used. The resulting multi-spectral images (\textit{panel B}) were then converted into 3 ``LMS'' channels using human cones sensitivity spectra and fed to the network. (\textit{Panel C}): the 4 illuminations R, G, Y and B were used exclusively in the evaluation. Note that while Y and B fall on the daylight locus, R and G have chromaticities different from the illuminations of the training set. Out of 1600, only 330 Munsell spectra were used.}
		\label{fig:methods}
\end{figure*}

The training was supervised with the learning objective of outputting the Munsell chip label for each image (i.e., the color of the object in each scene). Cross entropy was used as loss. Training took place for 90 epochs. As learning algorithm, we used the Adam optimizer \cite{kingma2015adam}, with a learning rate of 0.001, divided every 30 epochs by 10. We used a 40\% dropout as regularization for the input of the classification layer.


We trained separate models on the \textit{CC} and \textit{D65} datasets. Datasets were divided into training and validation. The test sets were generated separately, following a procedure described bellow. For each Munsell class, 28 images were selected for validation. These 28 images corresponded to the same 28 illuminations across Munsell class, randomly selected among our pool of 279 natural illuminations.
We can now see how our task requires our models to become color constant: in order for the models to achieve a high recognition accuracy on the \textit{CC} dataset, they would need to compensate for the chromatic shifts which are induced by the varying illuminations interacting with the lambertian surfaces. By extension, this means they would need to achieve some degree of color constancy.

Given that there are 2 datasets, \textit{CC} and \textit{D65}, 2 kind of training instances need to be distinguished: \textit{DeepCC} when trained on \textit{CC}, and \textit{Deep65} when trained on \textit{D65}. Due to several randomisation procedures implemented during training, 2 training instances of the same architecture trained on the same dataset will give slightly different results. To allow broader claims and a statistical analysis, we trained 10 instances of \textit{DeepCC} and \textit{Deep65} each. 

Each model was trained on one GeForce GTX 1080. Batchsize varied from architecture to architecture, but was maximized to fit the GPUs memory.  In the case of \textit{Deep}, the batchsize was 800 images.
All the code is available on Github \footnote{\url{https://github.com/AlbanFlachot/color\_constancy}}.

Other than the validation dataset, we devised other datasets to further evaluate our models. These evaluation datasets mimicked the typical experimental procedures for studying color constancy, consisting in removing or ambiguously modifying contextual cues to make the task more difficult \cite{witzel2018color, kraft2002surface}. They facilitated the identification of the diverse cues' relevance for the task, the testing of the model's robustness to scene modifications as well as the comparison with psychophysical studies.
These contextual modifications are: (1) removing the colored patches in the background. If the models use the constancy information transmitted by these patches, a drop in performance should follow; (2) placing the floating object in a background illuminated with a wrong illumination. If the models follow the information within the scene to estimate the illumination's color, then the resulting incorrect estimation should lead to a misclassification of the floating object's color; (3) removing the background entirely.

A detailed description of the evaluation datasets will follow in Sections \textit{Evaluation DeepCC and Deep65} and \textit{Impoverished visual scene}, sections where the results of these evaluations are presented.

\subsection{Metrics}

To assess the performance of \textit{DeepCC} and \textit{Deep65}, we used several measures of accuracy. Given that the task is the classification of Munsells, two are the standard top-1 and top-5 accuracies \cite{krizhevsky2012imagenet}: top-1 counts as hit when the correct Munsell is the one selected as most probable by the model; top-5 counts as hit when the correct Munsell is the 5 selected as most probable by the model. In addition, we defined the Muns$^3$ accuracy: counts as hit when the Munsell selected as most probable by the model is 1 Munsell away from the correct one (within a cube of side 3 in Munsell space centered around the correct Munsell).

Due to their discrete nature, however, top-1, top-5 and Muns$^3$ accuracies do not discriminate between cases when a model selected a Munsell just outside Muns$^3$ or when it was completely off. To correct this shortcoming, we converted the model's output into chromaticity coordinates. We did so by considering the Munsell chips chromaticities under the D65 illuminant in CIEL*a*b* space. We then defined the model's \textit{selected chromaticity} as the chromaticity of the Munsell selected by the model. The euclidean distance between the correct Munsell's chromaticity and the model's selected chromaticity now defines a continuous measure of the model's error. Following the literature \cite{CIELAB, weiss2017determinants}, we call this error $\Delta$E (with its 1976 definition). 

To further compare with the color constancy literature, we considered another measure called the \textit{Color Constancy Index} (CCI) \cite{Foster:2011, arend1986simultaneous, weiss2017determinants}. This measure has the benefit of taking into account the quantitative error of the model in color space ($\Delta$E) relative to chromaticity shift induced by the illumination. Consider that we present to the model an image showing a floating object under an illumination $I$ with the surface reflectance of a Munsell $M$. Consider now that the model recognizes the wrong Munsell $N$. Then the Color Constancy Index is defined as:

\begin{equation}\label{eq:CCI}
\begin{split}
CCI &= 1 - \frac{\left \vert C_I^N - C_I^M \right \vert}{\left \vert C_{D65}^{M} - C_I^M \right \vert},\\
&= 1 - \frac{ \Delta E }{\left \vert C_{D65}^{M} - C_I^M \right \vert}.
\end{split}
\end{equation}

where $C_M^I$ is the chromaticity of the Munsell $M$ under the illumination I, $C_M^{D65}$ is the chromaticity of the same Munsell chip but under the standard illumination D65 and $C_N^I$ the chromaticity of Munsell $N$ under the illumination $I$ and recognized by the model. If the model recognizes the correct Munsell, then the ratio in the formula is neutral and CCI would be equal to 1. However, if the model does not compensate for the illumination's shift in chromaticity and recognizes the wrong Munsell chip, CCI would be close to 0. Negative values of CCI indicate that the network chose the wrong Munsell for other reasons, beyond the chromaticity shifts induced by the illumination.

\section{DeepCC and Deep65 evaluation}

This section focuses on the evaluation of DeepCC and Deep65. Results for other architectures can be found in Section \textit{Standard and custum architectures}. 

We first present the results of training and validation for both DeepCC's and Deep65's instances. We then present thorough evaluations of the models using additional, custom datasets (description below).

We found that both DeepCC and Deep65 reached high top-1 accuracies on their respective validation datasets. DeepCC instances reached on average 76\% accuracy on the CC validation set, while Deep65 reached on average 86\% accuracy on the D65 validation set. These values clearly show that the 2 sets of networks learned how to solve their task, and are able to differentiate between 1600 different surface colors reasonably accurately (random performance would be 0.0625\%).  The higher performance of the Deep65 network also indicates, as expected, that the D65 task is inherently easier than when illumination is allowed to vary, and thus color constancy is required to perform the task.

In order to evaluate DeepCC in greater detail, as well as allowing some comparison with observations made in psychophysical studies, we generated another set of testing images, with settings closer to ``experimental" conditions.

\subsection{Methods} \label{ssec:evaluation}

To facilitate our analysis, an evaluation dataset was generated using a slightly different procedure than for the training sets. 
First, a subset of 330 Munsell chips was used, instead of the original set of 1600 (Cf. Figure \ref{fig:methods} \textit{panel C}). This subset was originally used for the World Color Survey and is now a standard for studies focusing on color naming \cite{berlin1991basic}. It is also widely used in studies related to color categories \cite{witzel2019misconceptions} and unique hues \cite{philipona2006color, flachot2016illuminant}. As such, they are an excellent basis for comparing our models with human judgments.

Second, we used 4 illuminations (Cf. Figure \ref{fig:methods} \textit{panel C}) equidistant to the CIEL*a*b* grey point by 10 $\Delta E$76 \cite{CIELAB} in the chromaticity plane. This procedure was inspired by experimental studies on illumination discrimination and estimation \cite{aston2019illumination}. Two, B and Y, lie on the daylight locus projected onto the chromatic plane. The other two, G and R, lie on the orthogonal direction which crosses the daylight locus at the grey point, and were outside of the distribution of illuminaations used during training. Their power spectra were generated with the principal components of natural daylight spectra defined by Judd et al. \cite{judd1964spectral}, which serve as the basis for the D series of the CIE standard illuminations. These illuminations were normalized such that their areas under curve were equalized, thus minimizing their difference in Lightness. 
For each Munsell of the 330 Munsell classes and each of the 4 illuminations, we generated 10 images for a total of $330\times4\times10=13200$ images

Note the fundamental difference between the validation sets employed earlier and the evaluation set defined now: while the validation datasets consisted of illuminations and 3D shapes the networks had never seen (to account for overfitting), these illuminations and shapes were still taken randomly from the same distributions as for the training set. The evaluation dataset, however, includes illuminations that are completely outside of the illumination distribution used at training time. As such, our evaluation procedure is in accordance the recommendations from the machine learning community and formally defined recently \cite{geirhos2020shortcut}:  using one \textit{independent and identically distributed} (i.i.d.) test set - our validation set - and another \textit{out of the distribution} (o.o.d) test set - this evaluation set.

Although the illumination spectra are different from the ones used during training and validation, the scene in which the floating objects are displayed is exactly the same. We therefore refer to this evaluation dataset as \textit{normal}. Because we are evaluating \textit{DeepCC} and \textit{Deep65}, each trained on different datasets, we distinguish between 2 conditions: \textit{CC} and \textit{D65}.

\begin{figure*}[t]
\centering
    \includegraphics[width=2\columnwidth]{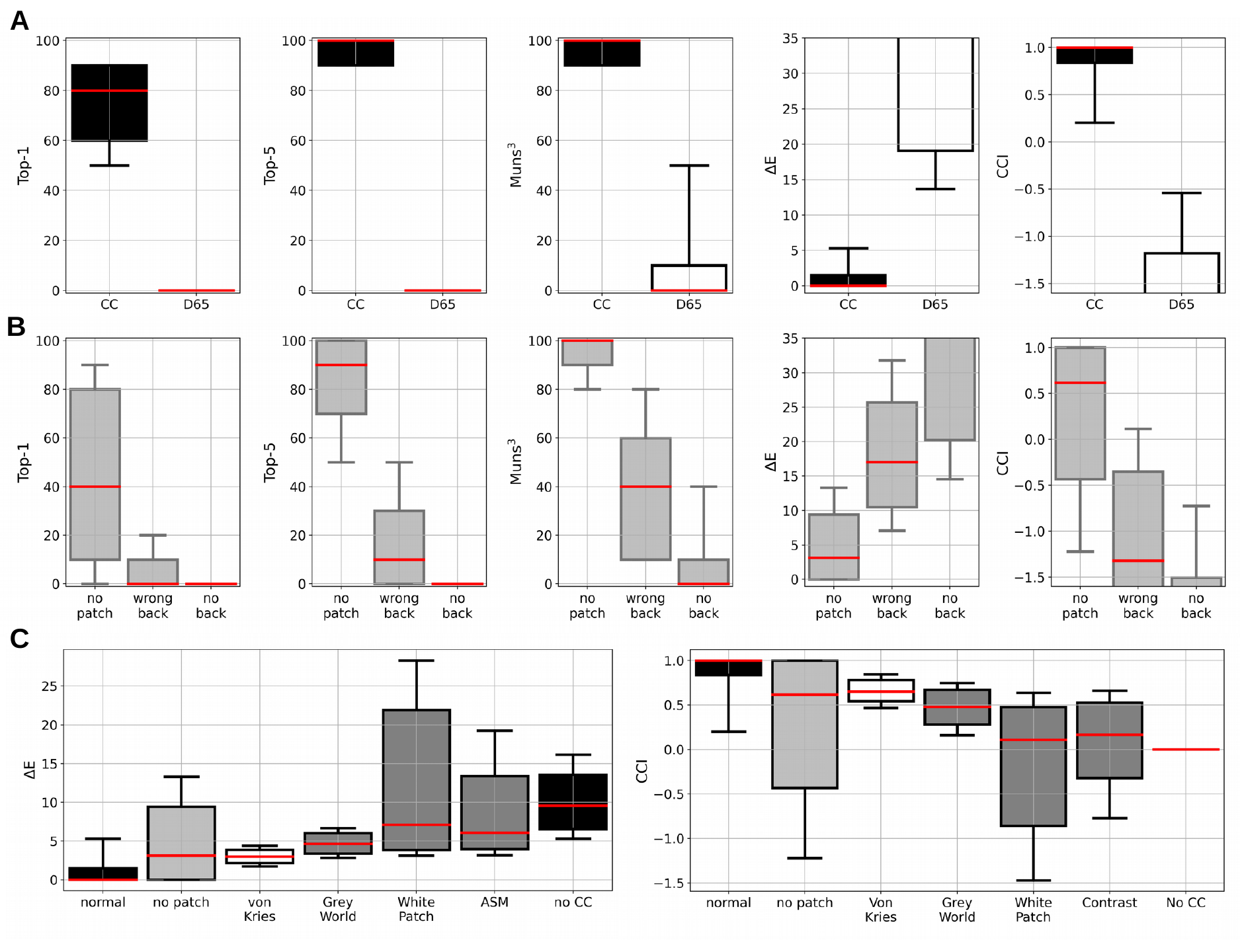}
\caption{DeepCC's evaluation results obtained for all measures and all conditions. Panel A: Performance under the \textit{CC} and \textit{D65} conditions (Section \textit{DeepCC and Deep65 evaluation}); Panel B: Performance of DeepCC under the \textit{no patch}, \textit{wrong back} and \textit{no back} (Section \textit{Impoverished visual scenes}); C: Performance of DeepCC compared to classical approaches to color constancy (Section \textit{Classical approaches}).} 
\label{fig:measures}
\end{figure*}

\subsection{Results}

Figure \ref{fig:measures} A shows the distributions obtained for each of our 5 metrics under the CC and D65 conditions. For the accuracies, we considered the distributions of values found for each Munsell class and illuminations (each point of the distribution is thus computed with 10 images). For $\Delta$E and CCI, we plot the distributions of values found for individual images.
Under the CC condition, we found median top-1, top-5 and Muns$^3$ accuracies of 80\%, 100\% and 100\% respectively across Munsell classes. The first quartiles are at 60\%, 90\% and 90\% respectively. This means that for the majority of Munsell classes, DeepCC selects the correct Munsell class in 4 out of 5 images, and when wrong it still selects a neighboring chip. This is confirmed by the distributions found for $\Delta$E and CCI, with median values of 0 and 1. 85\% of the images yielded less that 5 $\Delta$E error as indicated by the whiskers. As a comparison, note that the median $\Delta$E distance between adjacent chips is approximately 7.5. This means that when DeepCC instances selected the wrong chip, it tended to be a neighbour of the correct one, and a particularly close at that. 
Similarly, DeepCC showed a CCI higher than 0.83 in 75\% of cases. This CCI value of 0.83 takes its place among the higher ends of CCI measured in humans psychophysical experiments (cf. \cite{Foster:2011, witzel2018color} for reviews), thus indicating the supra-human performance of the model on this dataset. We also found a positive CCI value in more than 87\% of cases, evidence that DeepCC not only learned to discriminate between Munsell with high accuracy, but also learned to account for potential color shifts induced by the illumination.

Results were, however, very different for Deep65---the network trained using only a single illuminant, D65. We found median values of 0 in all 3 accuracies, meaning the 10 training instances of Deep65 rarely came close to selecting the right Munsell class. This is made clear with the distributions of the $\Delta$E and CCI measures. For the vast majority of the images, Deep65 exhibited errors of above 10 $\Delta$E and negative CCI, meaning that Deep65's error cannot be explained by the illumination change alone. This indicates that Deep65 lacks the ability to cope with illuminant deviations from the one it has been trained on, whereas DeepCC could generalize to novel illuminants beyond the 279 different illuminants it had been trained upon.


\subsection{Interim conclusion}

Results of this section clearly show that DeepCC did learn to accurately classify color surfaces under varying illumination. In doing so, it also learned to discount the illumination color, reaching a high degree of color constancy, even for illuminations outside of the gamut of illumination spectra used for training. Deep65, on the other end, performed very poorly on the 4 illuminations used for testing.

\section{Impoverished visual scenes: Performance with fewer cues} \label{sec:evaluation_impoverished}

We have seen that DeepCC achieved supra-human performance under normal condition on the devised evaluation dataset, thus achieving some degree of color constancy. A remaining question is what are the elements within the scene that DeepCC might use to compensate for illumination change: does it even consider the 6 constant color patches in the background? Given that there are inter-reflections between the floating object and the surrounding walls, is there any need for the model to use cues in the background at all? 

Computer graphics allow us to manipulate the scene elements as we wish. We thus devised new datasets to gain insights into which cues within the images DeepCC might use to achieve color constancy. Three manipulations were conducted: removing the constant patches in the background; showing a floating object illuminated by an illumination in a scene illuminated by another illumination; removing the context of the floating object entirely.

\subsection{Methods}

We generated 3 new image dataset to test DeepCC, in which some elements within the scene were removed or incongruously modified. These elements constituted cues that are known to be useful to humans for solving color constancy. Previous experiments\cite{kraft2002surface} have shown that increasing the color cues within a scene, in their case adding a Macbeth color checker, can increase color constancy for humans. Thus in one dataset, the  \textit{no patch} dataset, we removed the 6 constant patches located on the back wall. 
Other studies \cite{kraft1999mechanisms} showed that human color constancy is neutralized when the context surrounding the object of interest is manipulated incongruously. Thus in another dataset, \textit{wrong background}, the floating object, illuminated by one illumination, was cropped and placed in an empty scene illuminated by another illumination. In the final dataset, the floating object was cropped and placed in an empty space, thus without any context whatsoever, providing a baseline condition under which color constancy should be practically impossible (with the possible exception of cues provided by inter-reflections within the object). Note that for these last 2 conditions, human observers would be expected to be unable to solve the task.
Examples of images illustrating these conditions are shown in Figure \ref{fig:methods} \textit{D}.

\begin{figure}
\begin{tabular}{@{\hskip 0pt}c @{\hskip 5pt}}
     \includegraphics[width=1\columnwidth]{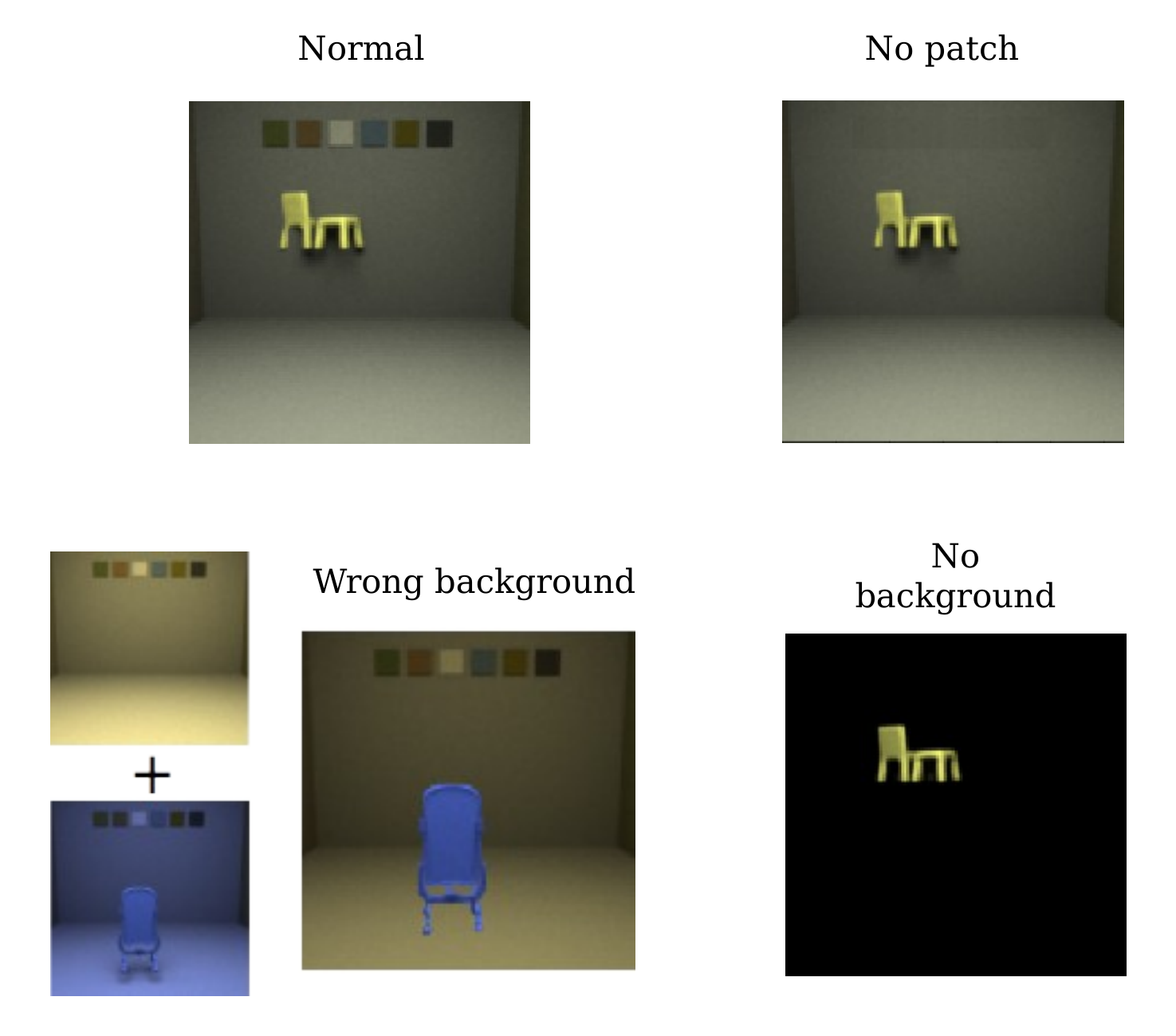}
\end{tabular}
\caption{Example for the 4 type of images we used during testing: normal, no patch (the colored patches in the background are removed), wrong background (the object in cropped and placed in room illuminated by another illumination) and no background (the object in floating in a dark room). } 
\label{fig:YBGRexample}
\end{figure}

\subsection{Results}

Results are shown in Figure \ref{fig:measures} B. Overall, DeepCC performed significantly worse in each of the 3 new conditions than in the normal condition, but sill better than Deep65 in the normal condition. Performance for the \textit{no patch} condition were on average still fairly high, indicating that the networks did not rely solely on the constant patches to perform the task. The three accuracy distributions include medians of 40\%, 90\% and 100\%. Muns$^3$ in particular shows a first quartile at 90\% accuracy, evidence that DeepCC was selecting for 3/4 of the Munsell classes a Munsell chips within the direct vicinity of the correct one under this condition. $\Delta$E and CCI measures lead to the same conclusions: median $\Delta$E is found at 3.3 and third quartile at 9.40, thus showing that in the large majority of cases, the model showed an error of the same magnitude as the inter-distance between Munsell chips in CIEL*a*b*. The analysis of the CCI distribution leads to the same conclusions: we found a median value of 0.62 but a first quartile at -0.43. This indicates that, while for most images DeepCC performs relatively well under the no patch condition (a CCI of 0.62 remains in the upper-half of CCI values reported in humans psychophysics), it is generally more difficult for the model to solve the task. To the extent that a significant proportion of images yield a negative CCI.

Interestingly, the reliance on the back patches was not equal across DeepCC's training instances. One instance saw its accuracy change by merely 20\%, while another experienced a drop of 41\%, twice as much. Refining our scene manipulations, we also looked at how the model's instances responded when masking one colored patch in the background at a time. Some patches appeared more critical than others: masking the red and yellow patches (2nd and 5th from the left) led to the largest loss in accuracy, with average losses of 9.9\% and 8.9\%. Masking the white and black patches (3rd and 6th from the left), however, had the least impact on the models performance, accounting for losses of 0.1\% and 4\% respectively on average. Individual differences were also confirmed. When masking the red patch, for example, one instance dropped by 22\% in accuracy, while another dropped only by 2.3\%. Some instances were also mainly affected by the red patch, other by the yellow patch.


These results are evidence that DeepCC indeed uses the information provided by the 6 constant colored patches in the back wall  -- particularly the chromatic ones. Nevertheless, the relatively high accuracies and CCI show that the model remained able to perform the task, albeit less successfuly.
The same cannot be said, however, when modifying the context entirely. In the \textit{wrong back} conditions DeepCC's median top-1 accuracy falls to 0, and the model's instances have median CCI values below 0. The top-5 and Muns$^3$ accuracies show, however, that the models still often selects a class of Munsell close to the correct one. 
These results are evidence that the model's instances heavily rely on the context surrounding the floating object to solve their task, similarly to humans. We would thus expect that the more different the background from the normal one, the more the model would fail. Conveniently, all illuminations of the training and validation datasets were used to generate the wrong background in a random fashion. Some images are thus bound to show a higher discrepancy between the illumination shining on the object and the illumination shining on the surroundings. We correlated the $\Delta$E error of the models with the $\Delta$E difference between the surrounding illumination and the object illumination, and indeed found a positive and significant correlation of 0.59.

Finally, we also have the results found in the \textit{no back} condition, where the background was entirely removed and replaced by pixel values of 0. In this case, as expected, the model's accuracy drops to the level of Deep65 tested on the CC evaluation dataset, described in the previous section.

\subsection{Interim conclusion}

Thanks to the controlled manipulation of the scene surrounding the floating object, we saw in this section that all DeepCC instances heavily rely on contextual cues to identify the object's Munsell surface and account for illumination change, in a similar fashion as humans \cite{kraft1999mechanisms, kraft2002surface}. Individual differences were observed between training instances, however, when the colored patches in the background were removed, some instances relying more on certain patches than others.  


\section{Comparison with classical approaches to color constancy} \label{sec:classic}

To further evaluate DeepCC, we compared its performance to the error expected with classical approaches to illumination estimation, coupled with the von Kries correction \cite{von1902chromatic}, standard in computer vision \cite{akbarinia2017colour, hu2017fc4}.

\subsection{Methods}

For comparison purposes we also computed on our test images of the CC normal condition, the errors expected from classical approaches to illumination estimation: \textit{Grey-World, White-Patch} \cite{retinax} and \textit{adaptive-surround modulation} (ASM) \cite{akbarinia2017colour}. All of these approaches are driven by low-level features (as opposed to learning): Grey-World makes the assumption that the world is on average ``grey" under a neutral illumination, and takes the average pixel value as an estimation of the illumination's color; White-Patch considers the brightest pixel as an estimation of the illumination; ASM assumes that image areas with high to middle spatial frequencies (typically edges) are most informative and computes the illumination by dynamically pooling a portion of the brightest pixels according the average image contrast. Each of these approaches delivers a single global triplet of values specifying the illuminant for a given image.

To enable a link from the global illumination estimations to our classification task of the floating object's surface color, we coupled these approaches with a global von Kries correction \cite{von1902chromatic}. This correction consisted in dividing each image pixel by the three illumination values estimated by each approach. For each resulting image, we then segmented the floating object, and estimated its chromaticity by considering the mean value of all its pixels. 
We then compared this estimated chromaticity with the chromaticity found for the exact same object, at the exact same position and orientation, but under a reference illumination. In this way, any difference between the estimated chromaticity and the reference chromaticity would be a consequence of the illumination estimation + correction only. As a reference, we used the computed chromaticity of the object rendered under the D65 illumination.

It is important to note that the von Kries method itself is of course only an approximation. It does not take into account inter-reflections within the scene for instance, or the loss of information resulting from the collapse of the spectral power onto cone activations \cite{worthey1986heuristic, flachot2016illuminant, Foster:2011}. We thus also estimated the error from the von Kries method by using the same procedure as for the other approaches for color constancy, but using the ground truth illumination instead of the estimated ones (perfect illumination estimator). Doing so, we also define an upper bound for these kind of approaches: no estimation illumination approach, as long as it is followed by von-Kries, can lead to a better result. Finally, we also computed the error obtained without compensating for the illumination at all. This would serve as an error estimate for a model that would perfectly segment the object of interest in the scene, but not compensate for the illumination (a perfect non color constant model).

\subsection{Results}

Figure \ref{fig:measures} C shows the distributions of $\Delta$E errors and CCI predicted from the classical approaches to color constancy, together with the results obtained under the normal and no patch conditions, described previously, for comparison purposes. Results obtained for the other conditions are not shown here, as we saw previously that the model's errors in these conditions could not be explained by the lack of color constancy alone, as the negative CCI indicate. 

We found median $\Delta$E values for all of the aforementioned approaches to be higher than for DeepCC under the \textit{normal} condition. Even the error merely induced by the von Kries adaptation (von Kries condition in the figure) leads to higher errors, with a median value of 2.9. This median value is in fact very similar to the median found for the \textit{no patch} condition, although slightly better. This is confirmed by the corresponding median CCI of 0.65.  Of the classical approaches, the Grey-World hypothesis proved to be the most accurate, with median values of 4.6 $\Delta$E and 0.48 CCI, slightly worse than for DeepCC on the \textit{no patch} condition. This suggests that not only did the DeepCC instances accurately identify the region of interest that is the object within the image and managed to accurately estimate the illumination, but they also accounted for the object's position with respect to the illumination and found a better correction strategy than a global von Kries. This is also a confirmation that DeepCC continues to perform well under the \textit{no patch} condition, albeit exhibiting a broader distribution than the von-Kries and Grey-World approaches.

We also evaluated the classical approaches under the \textit{wrong back} and \textit{no back} conditions. Similar to DeepCC's results, the $\Delta$E significantly increased in comparison to the \textit{normal} condition. Grey World remained the best algorithm among all classical approaches.

Grey-World's success compared to other approaches can be explained by the relative simplicity of the scene: a room with fairly neutral walls, with a single illumination. ASM would be expected to perform better using images of more complex scenes. The poor performance of the White-Patch approach for many images can be understood by the proximity of the object of interest to the camera: when a Munsell reflectance of high-value is applied to the object, the brightest pixels are likely to be found on the object itself, rather than on some other parts of the context.

\subsection{Interim conclusion}

Comparisons with classical approaches to color constancy show that under the \textit{normal} condition, DeepCC learned how to compensate for the illumination better than any of the classical approaches we tested. It even performed better than a hypothetical model provided with omniscient knowledge of the true illumination and compensating through the von Kries correction, the standard procedure for discounting in a scene the illumination after its estimation \cite{akbarinia2017colour}. 

\section{Effect of illumination on color constancy}

To test the DeepCC models, we used the 4 illuminations: Yellow (Y), Blue (B), Green (G) and Red (R) (see Figure \ref{fig:methods} \textit{B}). These were chosen to be equidistant to D65 in CIEL*a*b* space. Note however that even though none of these 4 illuminations were used during training, Y and B are expected to appear more ``familiar" to the models than the other two, due to the (natural) statistics of the training set. Natural light indeed varies much more along the Blue-Yellow direction than in the orthogonal direction, and this was reflected in our choice of natural illuminations for the training set. 

\begin{figure}
\begin{tabular}{@{\hskip 0pt}c @{\hskip 5pt}}
     \includegraphics[width=1\columnwidth]{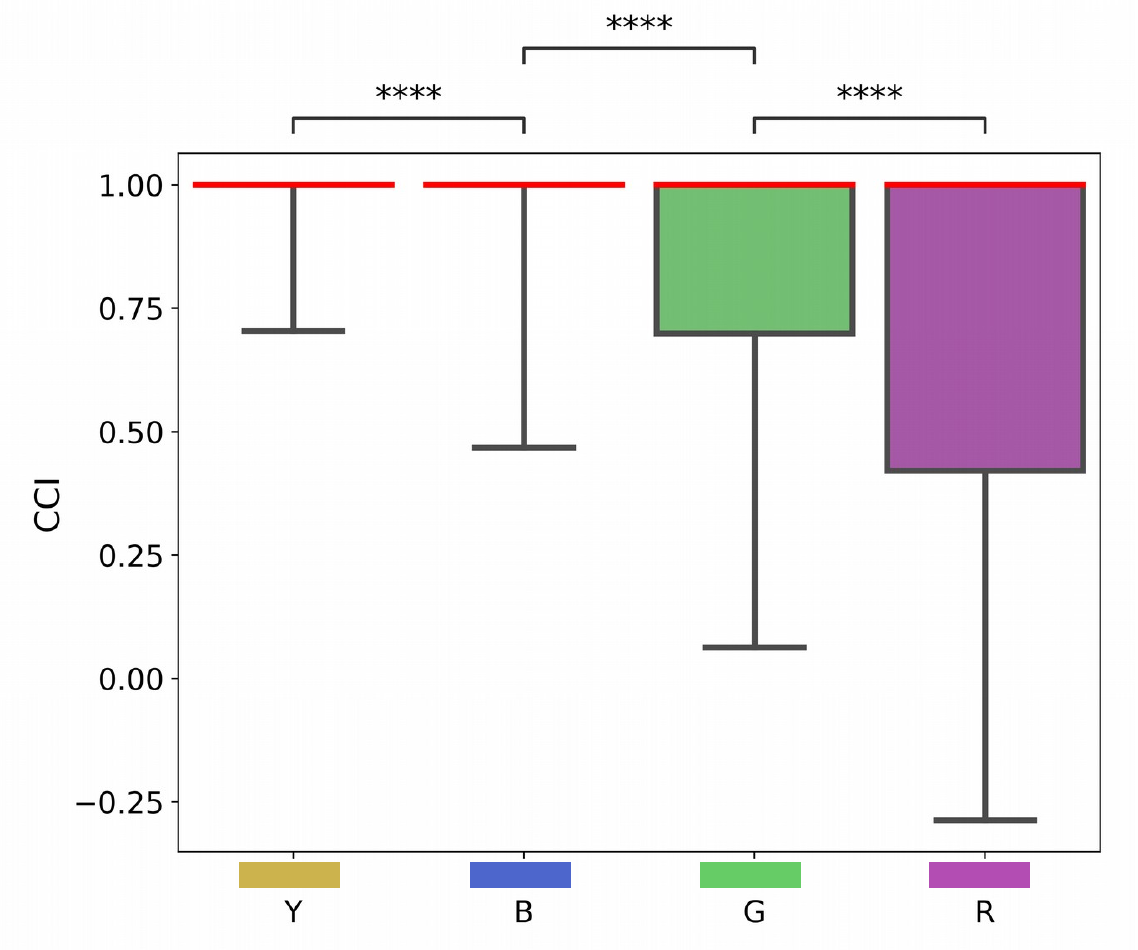}
\end{tabular}
\caption{Effect of the illumination on Color Constancy: Distributions of DeepCC's Color Constancy Index (CCI) under each of the 4 testing illuminations. Medians are in red. Statistical significance was computed applying pairwise t-tests with Bonferonni corrections.} 
\label{fig:YBGR}
\end{figure}

This anisotropy in the distribution of natural illuminations had consequences on the performance of our models, and in particular on their degree of color constancy. Figure \ref{fig:YBGR} shows the distributions of CCI across all test images in the \textit{normal} conditions, and across all instances. We observed a significant effect of the illumination on the CCI of our models: DeepCC models showed higher CCI for the ``familiar" illuminations (Yellow and Blue) than for the ``unfamiliar" illuminations (Green and Red). The highest degree of color constancy was found under the Yellow illumination with a mean CCI value of 0.96 while the lowest was found under the Red illumination with a mean CCI value of 0.82.

Results of Figure \ref{fig:YBGR} are very similar to observations made regarding the capacity of humans to perceive illumination changes \cite{pearce2014chromatic, aston2019illumination}. 
It was found that human observers were more sensitive to illumination changes happening along the green-red color direction compared to changes along the yellow-blue direction, meaning that they are less likely to perceive an illumination shift along the yellow-blue direction than along the green-red one. This suggests, the authors argue, that the human visual system compensates better for changes in the blue-yellow directions, which could have consequences for color constancy. 

\subsection{Interim conclusion}

Results in this section show a significant effect of the illumination on DeepCC's performance. Higher color constancy indices were observed for illuminations along the yellow-blue direction in CIEL*a*b* color space compared to illuminations falling onto the orthogonal direction. This difference is presumably explained by the model being more accustomed to variations along the daylight locus, the direction along which daylight and natural illuminations, such as the ones used for training, vary most. The parallel one can draw between our result and observations made in human psychophysics \cite{aston2019illumination} implies that the higher variation along the daylight locus may be a cause of similar consequences in humans.



\section{Color constancy throughout DeepCC}

There is uncertainty regarding where the neural mechanisms for color constancy would take place in the brain. Many studies emphasize on the importance of early mechanisms, such as cone adaptation \cite{lee1999horizontal}, or cells sensitive to chromatic contrasts between object and background in V1 \cite{wachtler2003representation}. Other have shown the lesions in macaque areas V4 also led to impaired color constancy \cite{wild1985primate} (see \cite{Foster:2011} for a review). In contrast to biological brains, deep neural networks like DeepCC allow access to the activations of every unit. Taking advantage of this, we added linear readouts to every layer of DeepCC in order to measure at which processing step color constancy emerges.

\subsection{Methods}

In order to apply the Color Constancy Index at different processing stages of DeepCC, we trained readout networks for each one of its 5 layers (3 convolutional and 2 fully connected). These linear probes \cite{alain2016understanding} consisted of very simple, fully connected linear models with 1600 kernels, 1 per Munsell class. For example, the readout network of DeepCC's first convolutional layer (RC1) takes as input the output of that layer and is trained on the same task as DeepCC, using the same dataset. The parameters of DeepCC's convolutional layer are not updated during this training iterations, only the weights of RC1. RC1 being fully connected and linear, no complex or non-linear operations are added and as such, RC1's performance is an indication of amount of information available in the first convolutional layer of DeepCC.

\subsection{Results}

\begin{figure}[t]
		\centering
			 	\includegraphics[width=1\columnwidth]{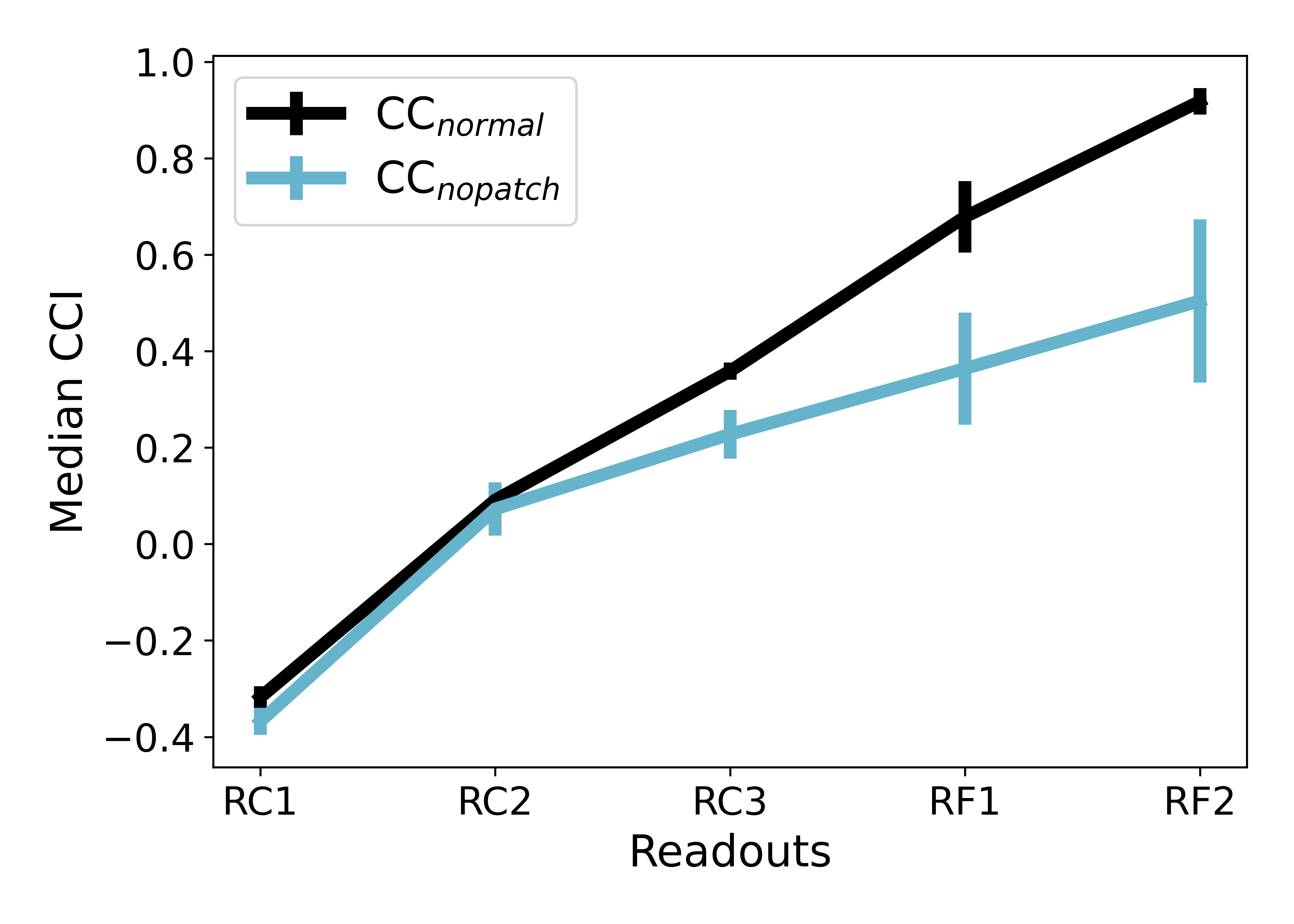}
		\caption{Color Constancy Index (CCI) for the 5 readout models tested with the \textit{normal} and \textit{no patch} image sets. Each readout take input from all units of the designated layer. By extension, the value of CCI reflects the degree of color constancy at the different layers of DeepCC.} 
		\label{fig:CCI_readouts}
\end{figure}

Figure \ref{fig:CCI_readouts} shows the average CCI obtained for DeepCC readout models. We named these readout models RC1, RC2, RC3 and RF1, RF2 corresponding to the convolutional layers 1, 2, 3 and the fully connected layers 1, 2 respectively. We trained 10 instances of each readout model, one for each instance of the original model. As shown in the plot, the readout models were tested under 2 conditions: CC$_{normal}$ (black) and CC$_{no patch}$ (cyan). Error bars are the standard deviation obtained across the 10 training instances. The color constancy index CCI gradually increases in the normal condition in an almost linear fashion across processing stages, consistently across the ten models. In the \textit{nopatch} condition, CCI follows the normal condition only up to RC2, at which point in continues increasing but at a much lower rate. The difference between the 2 conditions becomes significant from RC3 onwards. 
Error bars are also larger for the following layers, another indication of the large individual differences between training instances and observed in Section \textit{Impoverished visual scene}.

\subsection{Interim conclusion}

Contrary to many physiological studies emphasizing the early neural mechanisms for color constancy \cite{Foster:2011}, we found that color constancy seemed to increase steadily throughout DeepCC, both under the normal condition and the no patch condition. 


\section{Color representations in DeepCC}

We next performed a representational similarity analysis \cite{kriegeskorte2008representational} on unit activations within each layer to probe the models' internal representations of colors. We find that although the training objective treated each Munsell value as an entirely distinct class, the DeepCC networks nonetheless learnt similarity relationships between the colours that closely resemble their true embeddings in the CIEL*a*b* space.

\subsection{Methods}

To estimate the similarity between Munsell as seen by DeepCC, we computed the Representational Dissimilarity Matrices (RDMs)\cite{kriegeskorte2008representational} between Munsell classes for each layer in the DeepCC networks using correlation distance as a metric \cite{aguirre2007continuous}.
In turn, the RDMs were used as input to a Multi-Dimensional Scaling analysis (MDS) \cite{cox2008multidimensional}.


\subsection{Results}

\begin{figure*}[h]
		\centering
		\begin{tabular}{@{\hskip5pt}c}
		    \includegraphics[width=2\columnwidth]{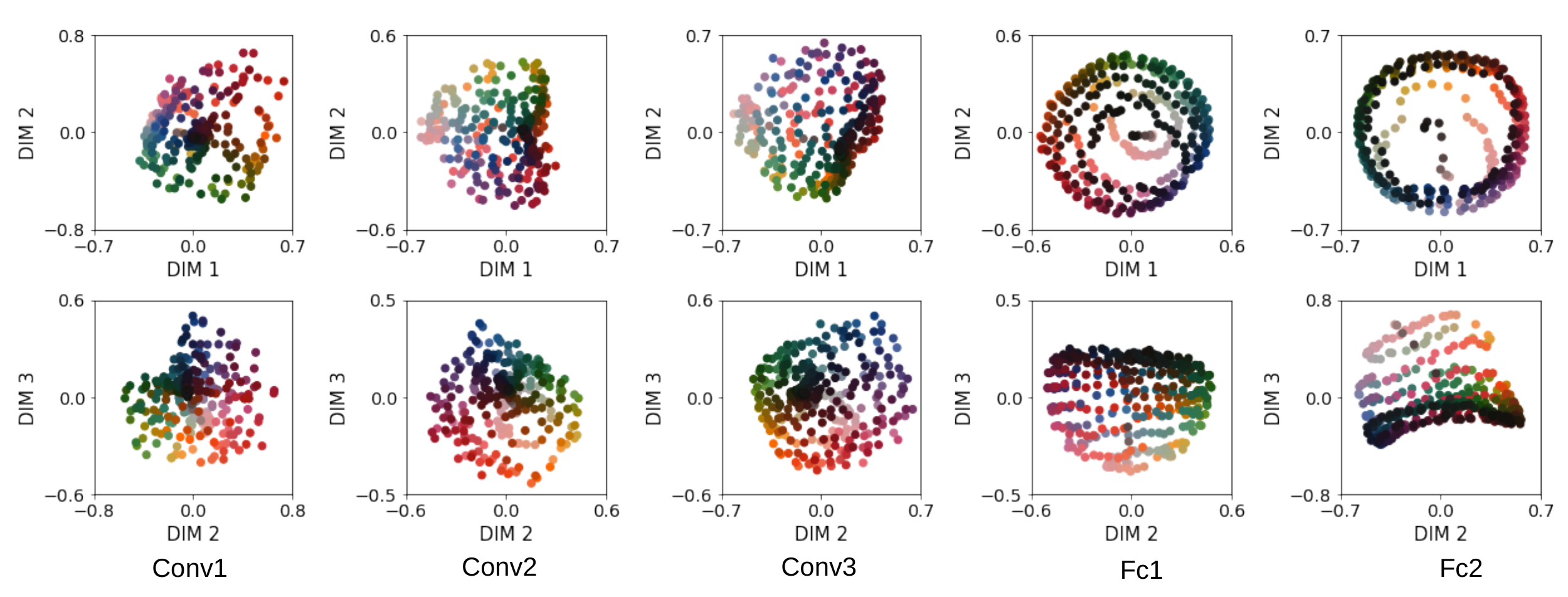}  \\
		\end{tabular}

		\caption{Results of a Multi-Dimensional scaling performed on the correlation of Munsell representations for different layers of DeepCC.}
		\label{fig:MDS}
\end{figure*}

We performed a MDS on the RDMs for each of the 5 layers of DeepCC.
Figure \ref{fig:MDS} shows 2D representations of the first 3 dimensions of the MDS results for each layer, tested under the \textit{normal} condition, and averaged across all 10 training instances. Each column corresponds to one layer. The upper row plots the first and second dimensions, the lower row the second and third. Colored dots correspond to Munsell chips and are displayed using their corresponding sRGB values.  
Interestingly, although the MDS could potentially yield a much larger number of dimensions, we found that except for the first layer, 3 dimensions were sufficient to explain more that 90\% of the variance, or dissimilarity between Munsell representations within the layers activations. This suggests that DeepCC mainly discriminate between Munsells according to 3 dimensions.
In addition, increasingly human-like color dimensions emerge in all layers: Munsells are separated according their hue, sometimes also their lightness. There is a progression in the way DeepCC represent Munsells: in early layers, many colors are clustered together in the center, rendering them less easily discriminable from one another. This changes in the last two layers, in which colors are more clearly separated from one another, and the dimensions become easier to understand. Particularly, in the first fully connected layer Fc1, each dimension seem to code for a standard color dimension: dimension 1  for "red-green" and dimension 2 for "yellow-blue". An almost perfect hue color circle, with a radius correlated with saturation is clearly visible. The 3rd dimension, shown in the bottom plot of the same column, separates Munsells according to their lightness. 

It is important to note that finding is nontrivial, and cannot be explained solely by the loss function we used. During training, the networks were never taught similarity relationships between Munsells. Rather, the error signal was the same whether the models wrongly selected a Munsell close to or far from the correct one in color space. Theoretically, a model could reach a high accuracy and not learn human-like similarities between the Munsell colors. And indeed, as reported below, other architectures trained and tested following the same procedures represent colors in a different manner.

\begin{figure}[t]
    \centering
    \begin{tabular}{@{\hskip5pt}c}
     \includegraphics[width=1\columnwidth]{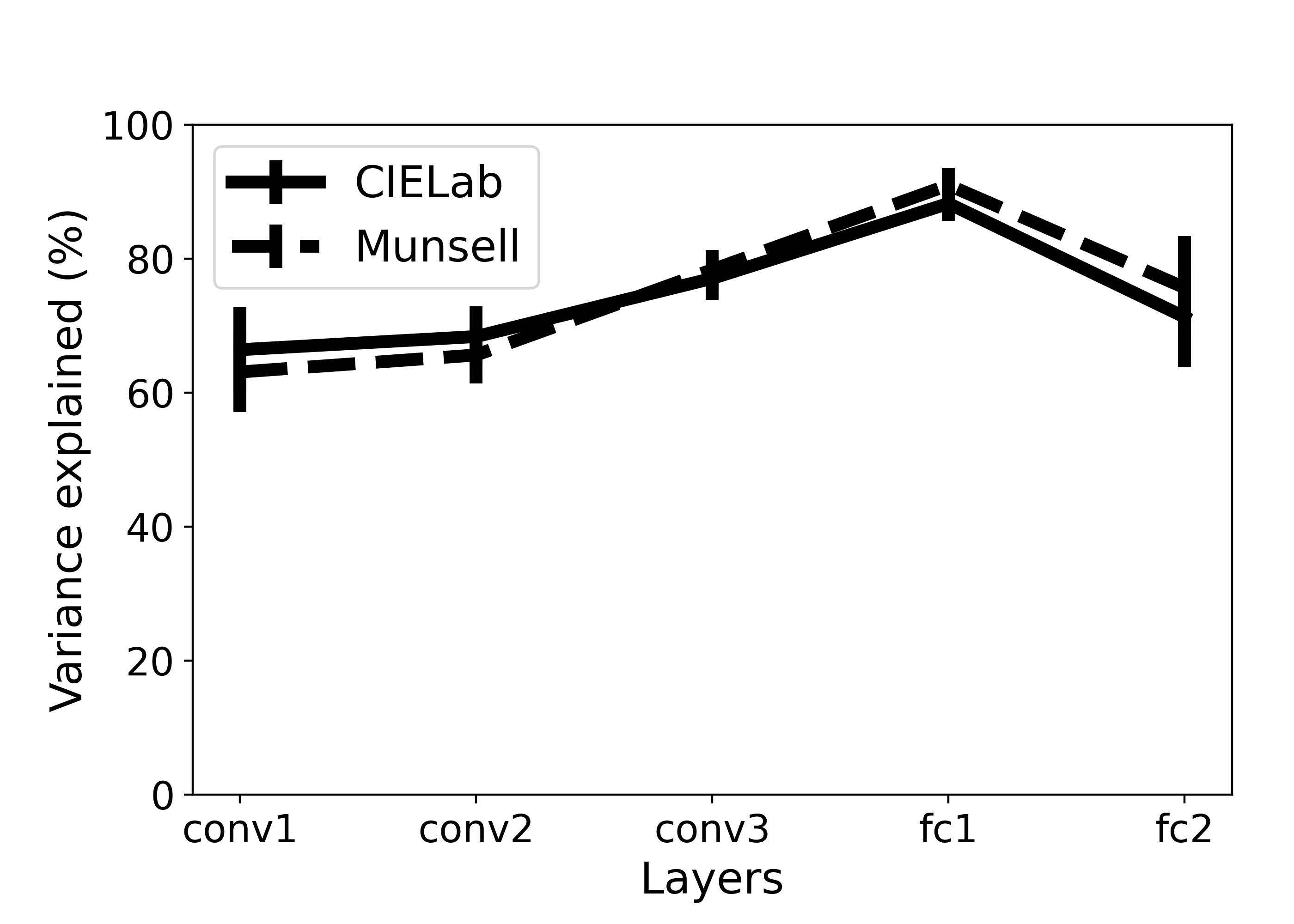}\\
       \end{tabular}
\caption{Results of the Procrustes analysis for DeepCC. The analysis is performed on the outcomes of the Multi-Dimensional Scaling at different layers using CIEL*a*b* and Munsell spaces as reference coordinates.} 
\label{fig:procrustes}
\end{figure}

We next sought to quantify the similarity---or difference---between Munsell representation in our models and their coordinates in a perceptual color space. To do this, we performed a Procrustes analysis \cite{gower1975generalized} to identify the rigid transformation that best mapped the coordinates obtained from the first three MDS dimensions, performed on each layer, to the corresponding coordinates in the Munsell and CIEL*a*b* color spaces.
The percentage of explained variance is an indication of the goodness of the mapping: the closer to 100\%, the better. As shown in Figure \ref{fig:procrustes} we find that in all layers, the variance explained is significantly higher than would occur by chance (95th percentile of 10.4\%), peaking at the FC1 layer which explains 91\% and 88\% of the variance for the Munsell and CIEL*a*b* color spaces respectively, indicating a very high similarity. The subsequent drop in FC2 likely reflects the demands of the objective function to deliver categorical outputs.

Qualitatively similar results were also obtained when using a Euclidean metric

 \subsection{Interim conclusion}

Similarly to the increasing CCI observed throughout the network in the previous section, the representational analysis also uncovered a progression in the way Munsell colors are represented within the model's layers. Visually, we could observe a progressive disentanglement of Munsell color values with increasing layer depth. More importantly, the representation of color also seemed progressively to resemble human perception, peaking at FC1, where there was a very high correspondence to the CIELa*b* perceptual color space. This was quantitatively confirmed using a similarity analysis, where it was found that the representational distances and dimensions between Munsells, in the penultimate layer in particular, matched very well the human perceptual distances and dimensions found empirically in psychophysical studies. 

\section{Standard and custom architectures}
\label{sec:StateArt}

We observed in the previous section that DeepCC seems to represent Munsell colors following color dimensions found empirically to be perceptually relevant for humans. Is this characteristic of DeepCC a general property of DNNs, or a special feature of this architecture?
To answer this question, we trained and evaluated several other standard deep learning architectures. 

\subsection{Methods}

\subsubsection{Architectures}

For the sake of comparison, we also trained three standard, high-performance deep learning models on the \textit{CC} dataset: VGG-11 \cite{simonyan2014very}, MobileNet \cite{howard2017mobilenets} and ResNet-50 \cite{he2016deep}. All of these architectures have specific features that make them significantly different from one another. 
These standard architectures, however, are relatively large and complex compared to the DeepCC architecture. While DeepCC only counts 676 kernels (outside of the classification layer's 1600) and 3.6M interconnections between unit, all three others count more than 13.000 kernels, the highest being ResNet-50 with almost 54.000. In order to allow some comparison with networks of a more similar size than DeepCC, we additionally devised another, shallower model. It consisted of a custom ResNet architecture, generated thanks to a ResNet bottleneck architecture generator (available in github \footnote{\url{https://github.com/ArashAkbarinia/kernelphysiology}}). To distinguish it to ResNet-50, we will call this architecture \textit{ResCC}. It has 3 layers, each with 3, 1 and 2 bottleneck blocks respectively. The first layer starts with 16 kernels, layer 2 with 32 and layer 3 with 64. Including the kernels within the bottleneck layers, it reaches 3424 kernels and 0.6M interconnections. Similarly to DeepCC, where we trained 10 instances, 6 independant training instances of ResCC were trained for further analysis.

\subsection{Results}


\begin{table*}[t!]
	\centering
	\begin{tabular}{@{\hskip5pt}c| @{\hskip 5pt} c@{\hskip 5pt} c c@{\hskip 5pt} c c@{\hskip 5pt} c c@{\hskip 5pt} c c @{\hskip5pt} c@{\hskip0pt}}
              Model         & \multicolumn{2}{c}{ MobileNet}    & \multicolumn{2}{c}{VGG-11}& \multicolumn{2}{c}{ResNet-50} & \multicolumn{2}{c}{ ResCC}    & \multicolumn{2}{c}{DeepCC (Ref ConvNet)}      \\
            Nb param & \multicolumn{2}{c}{4,3M} & \multicolumn{2}{c}{135,3M} & \multicolumn{2}{c}{29.8M} & \multicolumn{2}{c}{0.6M} & \multicolumn{2}{c}{3.6M} \\
			\hline 
              Condition     &   \textit{normal} &  \textit{no patch} &  \textit{normal}  &  \textit{no patch} &\textit{ normal} &     \textit{no patch} &   \textit{normal} &  \textit{no patch} &  \textit{normal} &  \textit{no patch} \\
            \hline
            Top-1    & 100 & 100 & 100 & 100 & 100 & 100 & 100 & 80 & 80 & 40 \\
            Top-5    & 100 & 100 & 100 & 100 & 100 & 100 & 100 & 100 & 100 & 90 \\
            Muns$^3$ & 100 & 100 & 100 & 100 & 100 & 100 & 100 & 100 & 100 & 100 \\
            $\Delta$E & 0.0 & 0.0 & 0.0 & 0.0 & 0.0 & 0.0 & 0.0 & 0.0 & 0.0 & 3.3 \\
            CCI      & 1.0 & 1.0 & 1.0 & 1.0 & 1.0 & 1.0 & 1.0 & 1.0 & 1.0 & 1.0 \\
			 \hline 
              Condition   &  \textit{wrong back} & \textit{no back} &  \textit{wrong back} &  \textit{no back} &  \textit{wrong back} &  \textit{no back} &  \textit{wrong back} &  \textit{no back} &  \textit{wrong back} &  \textit{no back} \\
            \hline
            Top-1    & 0.0 & 0.0 & 0.0 & 0.0 & 0.0 & 0.0 & 0.0 & 0.0 & 0.0 & 0.0 \\ 
            Top-5    & 0.0 & 0.0 & 0.0 & 0.0 & 0.0 & 0.0 & 0.0 & 0.0 & 10 & 0.0 \\
            Muns$^3$ & 40 &	0.0 & 40 & 0.0 &	40 &	0.0 &	40 &	0.0 &	40 &	0.0\\
            $\Delta$E & 18.9 &	41.1 &	16.1 &	61.9 &	16.5 &	57.11 &	17.2 &	63 &	17.0 &	36.7 \\
            CCI      & -1.31 &	-5.2 &	-1.23 &	-7.75 &	-1.24 &	-6.23 &	-1.35 &	-7.03 &	-1.32 &	-3.9 \\

	\end{tabular}
	\caption{Median values found for all measures and all models under the normal, no patch, wrong back and no back conditions.}
	\label{tab:accuracies}
\end{table*}

We evaluated each one of the DNN architectures on the \textit{normal}, \textit{no patch}, \textit{wrong back} and \textit{no back} conditions. Here, for the sake of simplicity, we chose to show only a summary of the results with a table with the distributions' medians. 
Table \ref{tab:accuracies} shows the medians obtained for all architectures under those conditions, with the results obtained for DeepCC as a reminder on the last column. MobileNet, VGG-net, ResNet-50 and ResCC all showed very high performance under the \textit{normal} and \textit{no patch} conditions, higher than DeepCC. Importantly, there was almost no difference between the 2 conditions for these networks, meaning that they make little to no use of the patches in the background.
All models have shown a significant loss in accuracy for the \textit{wrong back} and \textit{no back} conditions, suggesting that all tested models rely heavily on cues in the background to perform their task.


Up to now, standard networks and ResCC essentially shared the same characteristics as DeepCC: while they outperformed the classical approaches to color constancy, such as Grey-World (cf. Section (Comparison with classical approaches)) under the \textit{normal} condition, they failed under the \textit{wrong back} and \textit{no back} conditions (Cf. Figure \ref{fig:measures}), as indeed essentially any observer would.
Additionally, we found they also show a significant effect of the illumination on the Color Constancy Index, with higher performance for the Yellow and Blue illuminations than for the green and Red illuminations. 

However, when it came to the analysis of Munsell representations within its latent layers, they all exhibited a very different picture to DeepCC: Munsells did not appear to be differentiated following human-like color dimensions. Like in the previous section, we performed a Multi-Dimensional Scaling on the RDMs found for each layer of each architecture, followed by a Procrustes analysis using CIEL*a*b* as a reference space. Across all architectures, the highest percentage of explained variance resulting from the Procrustes analysis was 53\%. It was obtained for the VGG-11 architecture's 4th layer, and stands way below the 88\% explained variance of DeepCC's pernultimate layer. 

\begin{figure}[t]
    \centering
    \begin{tabular}{@{\hskip0pt}c}
     \includegraphics[width=1\columnwidth]{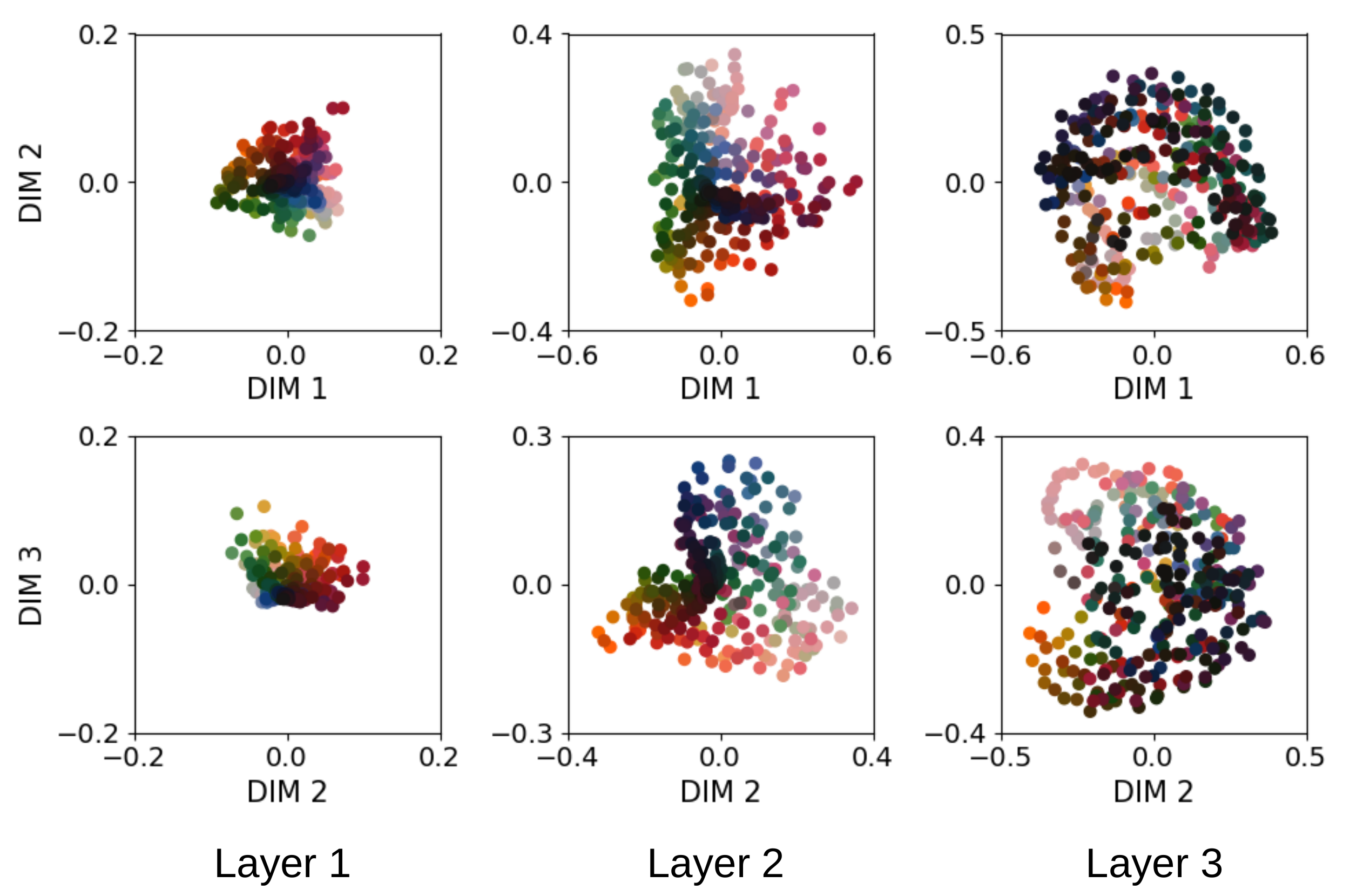}
       \end{tabular}
\caption{ Results of a Multi-Dimensional scaling performed on the correlation distance of Munsell representations for different layers of ResCC. Compared to DeepCC (Cf. Figure \ref{fig:MDS}), ResCC does not seem to classify Munsells following the same dimensions as those defined by human perception. Particularly in layer 3.} 
\label{fig:MDS_ResCC}
\end{figure}

As an example, we show in Figure \ref{fig:MDS_ResCC} the results of the MDS analysis averaged over the ResCC instances. We can observe that none of the three layers visibly separate Munsell colors human-like perceptual dimensions like Hue, Lightness or Chroma. This is particularly true for layer 3. For this last layer, the first 3 dimensions of the MDS account for only 54\% of the dissimilarity between Munsell representations, meaning that Munsell discrimination took place in a space with more than 3 dimensions.

\begin{figure}[t]
    \centering
    \begin{tabular}{@{\hskip0pt}c}
     \includegraphics[width=1\columnwidth]{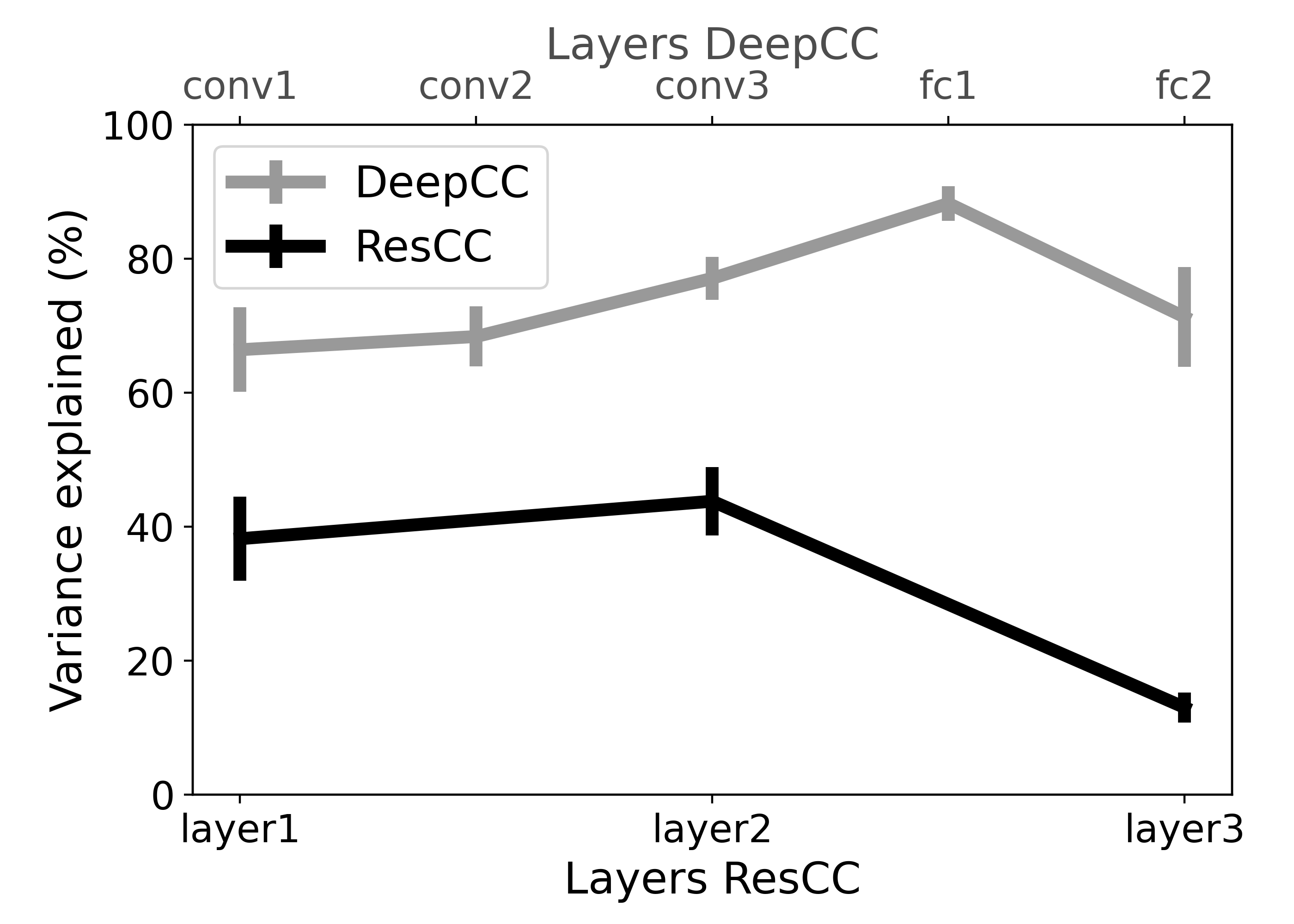}\\
       \end{tabular}
\caption{Results of the Procrustes analysis for both DeepCC (grey) and ResCC (black). The analysis was performed on the outcomes of the Multi-Dimensional Scaling at different layers using CIEL*a*b* spaces as reference coordinates. The variance explained for ResCC was consistently lower than for DeepCC throughout its layers, meaning that ResCC discriminate Munsells following color dimensions dissimilar to those defined by human color perception.} 
\label{fig:procrustes_all}
\end{figure}

This observation is further confirmed by Figure \ref{fig:procrustes_all}, which plots the outcome of the Procrustes analysis on the result of the Multi-dimensional scaling, using CIEL*a*b* as a reference space. As a reminder, the results of the same analysis for DeepCC are reproduced again in grey. The variance explained by the best fit for mapping Munsell representations in ResCC layers onto the CIEL*a*b* coordinates was always lower than for DeepCC, meaning that ResCC discriminate Munsells following color dimensions dissimilar to those defined by human color perception, contrary to DeepCC.

\subsection{Interim conclusion}

The results of our comparisons with other architectures show that if performance was our only goal, many architectures other than \textit{Deep} could have been used to solve the Munsell classification task, and indeed achieved superior performance. 
The similarity analysis we used, however, showed that other architectures, such as ResCC, seemingly differentiates between Munsell colors according to features (i.e., color dimensions) very different to those empirically found for human perception, contrary to DeepCC.

This last observation emphasizes the need for careful examination when it comes to selecting a DNN architecture for a given task. While at first sight, ResCC might have seemed a better choice for our tasks (highest performance and few parameters), the analysis of the Munsell representations shows that DeepCC presents characteristics more similar to human color discrimination. This last point suggests that DeepCC is thus potentially a better candidate for modeling human discrimination of Munsell color surfaces. It also emphasizes the need to develop further methods and strategies to analyse and understand the features learned by different architectures.

\section{Discussion} \label{sec:discussion}

\subsection{Deep Neural Networks for biological color vision}

We define and study Deep Neural Networks trained for color classification, a task which requires color constancy for good performance. We find that as a result of this training, the deep neural network models became similar to humans in several respects: They learned color constancy and were able to generalize to illuminations out of their training distribution. They used the contextual information like the colored patches in the background to solve their task. Furthermore, they developed a similar bias towards daylight-locus illuminations as observed in humans for estimating illumination changes.

In addition, our analysis on the inner representations of surface colors showed that DeepCC learned to discriminate colored surfaces following dimensions similar to the CIE-Lab space, which is based on human perception. This similarity seems to be the exception rather than the rule, as other architectures like ResCC, represented color in a different way. This difference is particularly interesting because the other architectures were more capable at solving the training task, reaching higher accuracies than ConvNets, and indeed achieving superior color constancy. The observation that one architecture learned human-like features and not the other hints at architectural influences shaping human color perception. Understanding these architectural influences better may help us understand human color vision in the future.


\subsection{Color discrimination and color constancy in Deep Neural Networks}

We have just discussed the usefulness of considering deep learning models for color discrimination and color constancy. We would like to argue also that the reciprocal is also true: color discrimination and color constancy are useful tasks for understanding and studying deep neural networks. The relative simplicity of color tasks allows for the use of simpler---thus easier to study---models than the state-of-the-art models for object recognition. The large number of studies on human color vision \cite{witzel2018color} make it possible to devise many out-of-distribution \cite{geirhos2020shortcut} testing conditions, like in this article, to evaluate the generalisability of our models. For all these reasons, color discrimination and color constancy are suitable tasks to study deep learning models.

\subsection{3D-rendered dataset for color constancy}

Unfortunately, large datasets consisting of numerous photographs of real, complex scenes with controlled conditions suitable to train deep neural networks from scratch on color constancy tasks do not yet exist. The popular ImageNet \cite{deng2009imagenet}, for instance, consists of millions of natural images but taken from non-calibrated cameras, presumably white-balanced. The ColorChecker dataset \cite{gehler2008bayesian} has the opposite characteristic: it presents precise and well calibrated complex images, but less than 1000 of them. Large hyperspectral datasets of natural scenes at different times of the day would be optimal of course, but the difficulty of controlled hyperspectral captures is such that most datasets count a few hundreds of images at most \cite{vazquez2009color, nascimento2016spatial}. 

As shown in this study, a spectrally-3D-rendered dataset thus seems like a good alternative: it allows a perfect control of the color conditions (illuminations, surfaces) with a certain realism (accurate physical models of lights and materials, spectral data). Object shapes, positions and material can be freely manipulated, and ground truth can be specified either using the labels of the color surfaces or through their rendered chromaticity under a reference illuminations like D65. Finally, it allows an easy manipulation of contextual cues, to test which ones are used by the respective models. While shown here for color discrimination and color constancy, other studies have shown the great potential of combining Deep Learning and computer graphics for studying other visual functions, like the perception of optic-flow \cite{dosovitskiy2015FlowNet} and the perception of liquids \cite{van2020visual}

Some challenges remain, however, such as the efficient creation of convincing outdoor scenes. It is possible that reproducing the statistics of more complex, naturalistic scenes would contribute towards greater robustness of DNNs to scene changes, and perhaps allow the emergence of higher features of color vision, such as color categories \cite{witzel2018color, parraga2016nice}.

\subsection{Implications for color constancy in general}

Our results have several implications for color constancy in general, independent of whether we believe that DNNs are a good model of human color constancy.  First, we  trained networks to extract the surface color more accurately than a perfect global von Kries correction. This implies that a global illumination correction is not the optimal solution to the color constancy problem, even in a situation with a single illumination color. This may guide future computer vision and image processing work that aims to extract object properties rather than color-correcting images. Second, we confirm earlier suspicions that the prior distribution over illuminations causes the better performance of humans along the daylight axis, as employing a naturalistic range of illuminations was sufficient to cause our networks to have this bias as well. Third, our finding that network architectures like ResCC can achieve outstanding color constancy performance despite not reproducing human perceptual color similarity representations suggests that these representations are not necessary for color constancy. Although perceptual color spaces presumably have many advantages for human color vision, our findings do not support the notion that they are specifically optimised for color constancy ---at least in the class of images we investigated. An interesting direction for future research would be to train networks explicitly on perceptual color representations and test how this improves performance at other tasks.  This would potentially provide answers to the teleological question of why human color space is shaped as it is \cite{dicarlo2012does}.

\section{Conclusion}

In this study, we approached color constancy as a surface reflectance discrimination task under varying illuminations, which closely mimics what humans do on daily basis. We modelled this using deep neural networks. This is a novel methodology in computational modeling of color constancy. We solved the need of a large dataset to train such models while safeguarding some naturalism and control over the stimuli by generating our images using a spectral 3D renderer. We trained several different architectures to explore the generalisability of our findings. 
We then devised a set of testing conditions to thoroughly evaluate our models and compare them to human behavioural studies. We found that similarly to humans, all models heavily relied on contextual cues to solve color constancy and show the same bias towards illuminations along the daylight locus as humans. However, a similarity analysis on the activations patterns within the deep latent layers of the trained models showed significant differences in the way they represented color surfaces. Only one convolutional network, DeepCC, learned to discriminate colored surfaces following very similar dimensions to those used by humans. This suggests that in computational models of human color constancy, the highest performance alone might not be the best metric to measure fidelity of a model to human color representations. This is in line with reports in object classification, where lower performance networks may better correlate with human brain recordings and behavioral measurements \cite{kubilius2019brain, geirhos2020beyond}.

\section{Acknowledgments}

We would like to thank our colleagues in Giessen for their support and useful discussions. In particular, we thank Guido Maiello for his statistical advice, Filipp Schmidt for making the meshes dataset accessible, Robert Ennis  and Florian Bayer for their discussions and software support. We also would like to thank Raquel Gil for her mathematical expertise and Christoph Witzel for his color expertise. In T\"ubingen we are grateful to Bernhard Lang for helping us to get started with spectral renderings and Uli Wannek for technical support. 

This work was funded by the Deutsche Forschungsgemeinschaft (DFG,
German Research Foundation) – project number 222641018
– SFB/TRR 135 TP C2. FAW was funded, in part, by the Deutsche Forschungsgemeinschaft (DFG, German Research Foundation) under Germany’s Excellence Strategy – EXC number 2064/1 – Project number 390727645.

\bibliography{refs}
\bibliographystyle{jovcite}

\end{document}